\def\year{2019}\relax
\documentclass[letterpaper]{article} 
\usepackage{aaai19}  
\usepackage{times}  
\usepackage{helvet}  
\usepackage{courier}  
\usepackage{url}  
\usepackage{graphicx}  
\usepackage{multirow}
\usepackage{subfig}
\usepackage{caption}
\usepackage{amsmath}
\begin{document}

%
\title{Scene Text Recognition from Two-Dimensional Perspective}
\author{\textsuperscript{1}Minghui Liao\thanks{Authors contribute equally.}, \textsuperscript{2}Jian Zhang\footnotemark[\value{footnote}]
, \textsuperscript{2}Zhaoyi Wan\footnotemark[\value{footnote}], \textsuperscript{2}Fengming Xie, \AND \textsuperscript{2}Jiajun Liang, \textsuperscript{1}Pengyuan Lyu, \textsuperscript{2}Cong Yao, \textsuperscript{1}Xiang Bai\thanks{Corresponding author.}\\
\textsuperscript{1}Huazhong University of Science and Technology,  
\textsuperscript{2}Megvii (Face++)\\ 
mhliao@hust.edu.cn, buaacszj@qq.com, i@wanzy.me, beautifeng@gmail.com,\\ liangjiajun@megvii.com, lvpyuan@gmail.com, yaocong2010@gmail.com, xbai@hust.edu.cn}
\nocopyright
\maketitle
\begin{abstract}
Inspired by speech recognition, recent state-of-the-art algorithms mostly consider scene text recognition as a sequence prediction problem. Though achieving excellent performance, these methods usually neglect an important fact that text in images are actually distributed in two-dimensional space. It is a nature quite different from that of speech, which is essentially a one-dimensional signal. In principle, directly compressing features of text into a one-dimensional form may lose useful information and introduce extra noise. In this paper, we approach scene text recognition from a two-dimensional perspective. A simple yet effective model, called Character Attention Fully Convolutional Network (CA-FCN), is devised for recognizing the text of arbitrary shapes. Scene text recognition is realized with a semantic segmentation network, where an attention mechanism for characters is adopted. Combined with a word formation module, CA-FCN can simultaneously recognize the script and predict the position of each character. Experiments demonstrate that the proposed algorithm outperforms previous methods on both regular and irregular text datasets. Moreover, it is proven to be more robust to imprecise localizations in the text detection phase, which are very common in practice.
\end{abstract}

\section{Introduction}
Scene text recognition has been an active research field in computer vision because it is a critical element of a lot of real-world applications, such as street sign
reading in the driverless vehicle, human computer interaction, assistive technologies for the blind and guide board recognition~\cite{DBLP:conf/eccv/RongYT16,ZhuLYL18}. As compared to the maturity of document recognition, scene text recognition is still a challenging task due to large variations in text shapes, fonts, colors, backgrounds, etc.

\begin{figure}[!htbp]
\begin{center}
\captionsetup[subfigure]{justification=centering}
    \centering
\subfloat[\label{subfig-1:introduction}]{%
       \includegraphics[width=0.2\textwidth]{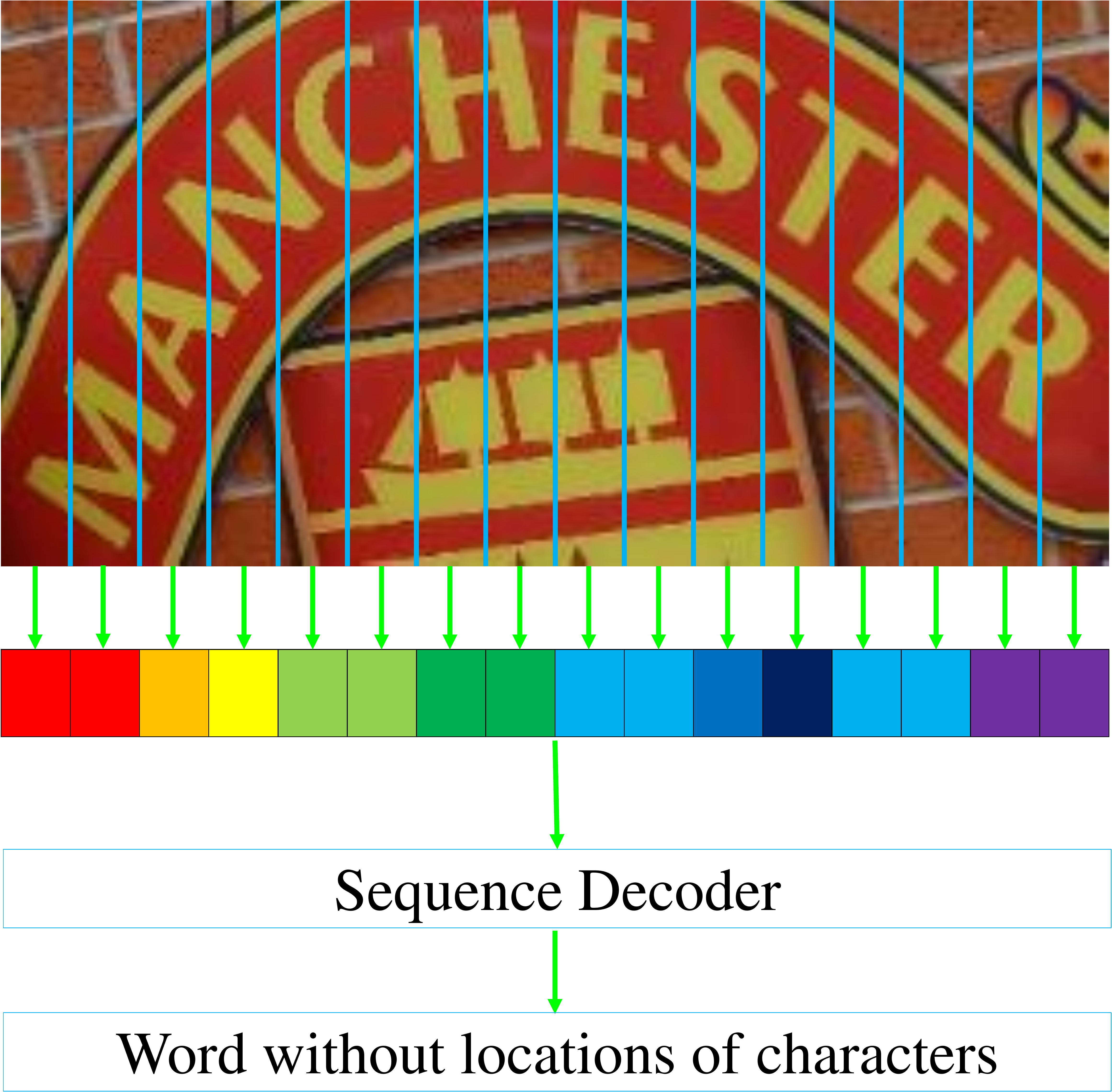}
     }
\subfloat[\label{subfig-2:introduction}]{%
       \includegraphics[width=0.2\textwidth]{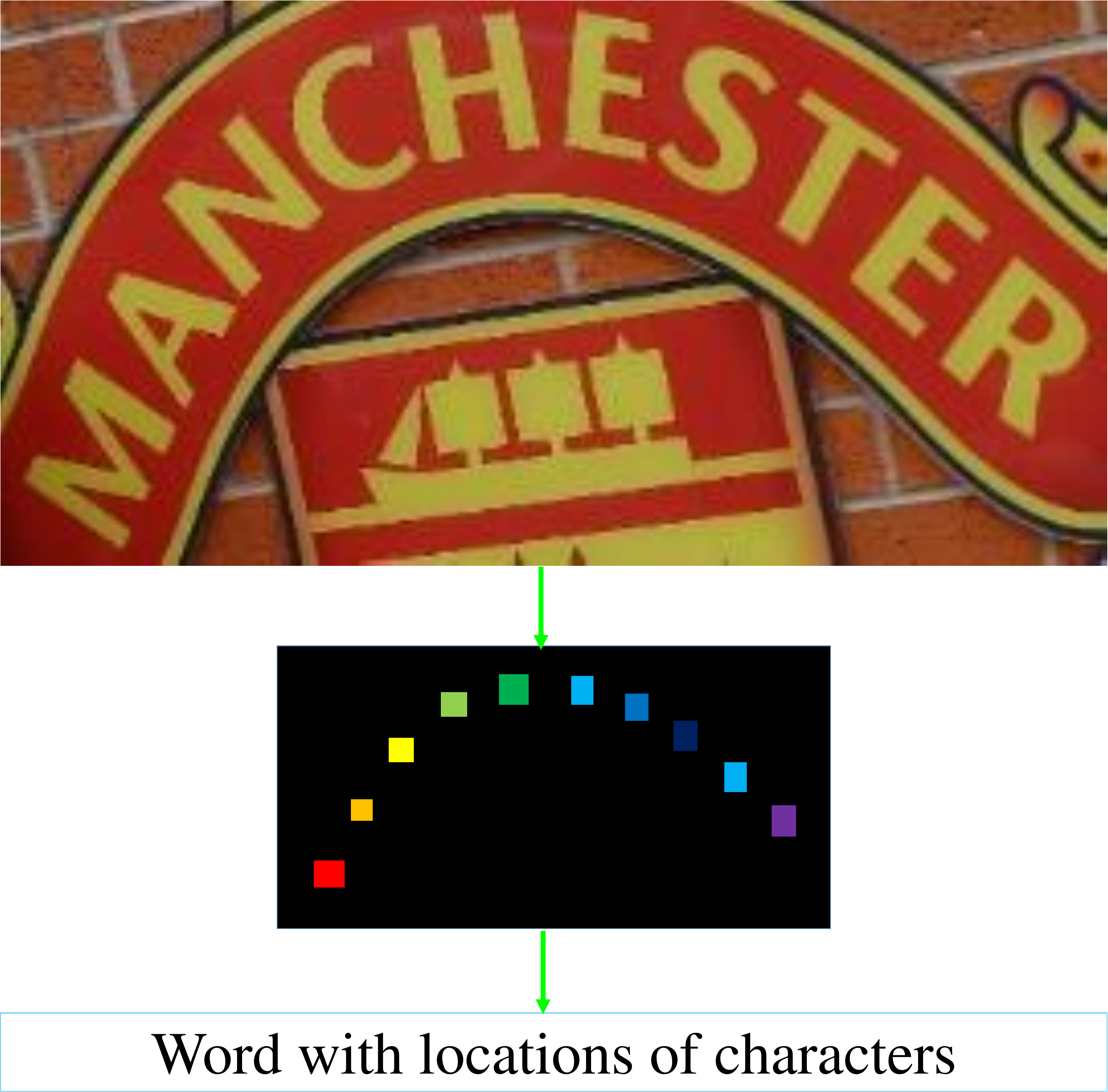}
     }
\end{center}
\caption{Illustration of text recognition in one-dimensional and two-dimensional spaces. (a) shows the recognition procedures of sequence-based methods. (b) presents the proposed segmentation-based method. Different colors mean different character classes.}
\label{fig:introduction}
\end{figure}

Most of the recent works~\cite{DBLP:journals/pami/ShiBY17,DBLP:conf/cvpr/ShiWLYB16,zhu2016scene} convert scene text recognition into sequence recognition, which hugely simplifies the problem and leads to great performance on regular text. As shown in Fig.~\ref{subfig-1:introduction}, they firstly encode the input image into a feature sequence and then apply decoders such as RNN~\cite{hochreiter1997long} and CTC~\cite{ctc} to decode the target sequence. These methods produce good results when the text in the image is horizontal or nearly horizontal. However, different from speech, text in scene images is essentially distributed in a two-dimensional space. For example, the distribution of the characters can be scattered, in arbitrary orientations, and even in curve shapes, as shown in Fig.~\ref{fig:introduction}. In these cases, roughly encoding the images into one-dimensional sequences may lose key information or bring undesired noises. 
\cite{DBLP:conf/cvpr/ShiWLYB16} tried to alleviate this problem by adopting a Spatial Transform Network (STN)~\cite{stn} to rectify the shape of the text. Nevertheless, \cite{DBLP:conf/cvpr/ShiWLYB16} still used a sequence-based model, so the effect of the rectification is limited.

\begin{figure*}[ht]
\centering
\includegraphics[width=0.85\linewidth]{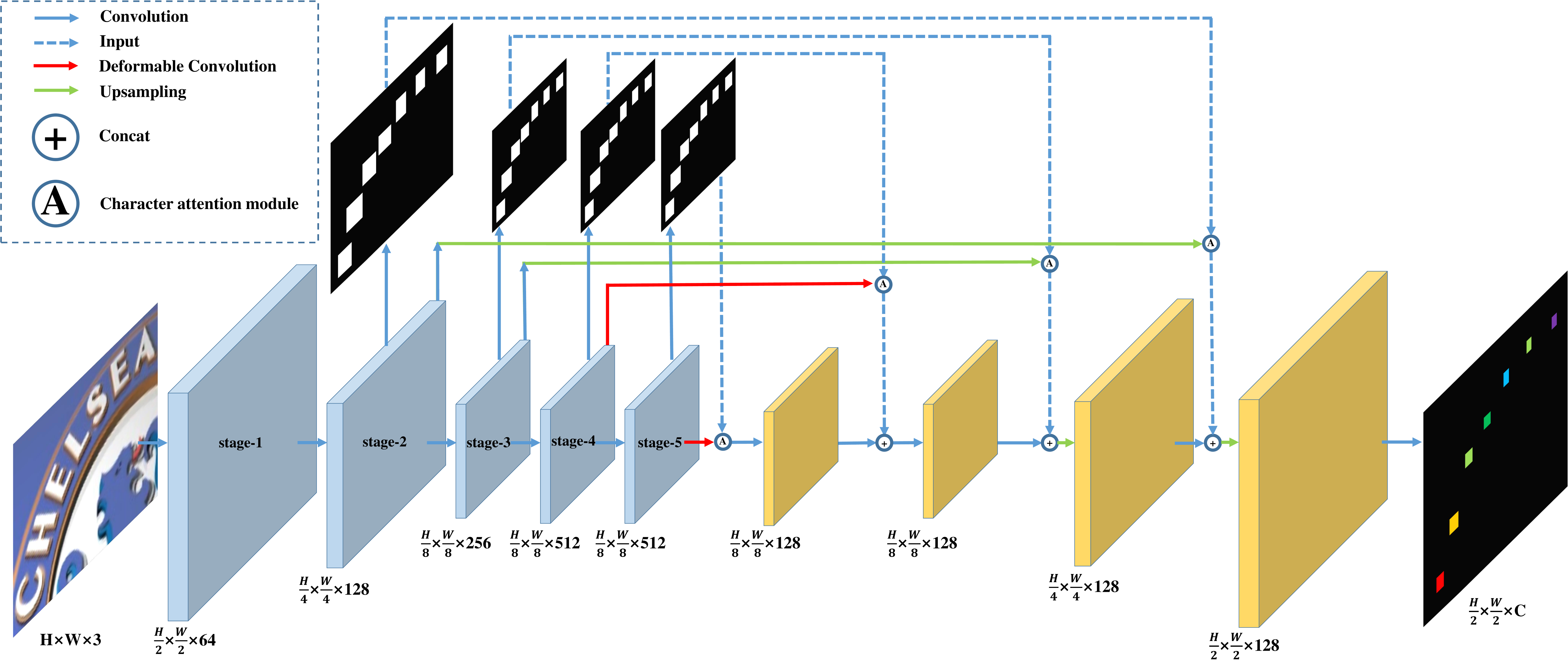}
\caption{Illustration of the CA-FCN. The blue feature maps in the left are inherited from the VGG-16 backbone; The yellow feature maps in the right are extra layers. H, W mean the height and width of the input image; C is the number of classes.}
\label{fig:network}
\end{figure*}

As discussed above, the limitations of sequence-based methods are mainly caused by the difference between the one-dimensional distribution of feature sequences and the two-dimensional distribution of text in scene images. To overcome these limitations, we tackle the scene text recognition problem in a new and natural perspective. We propose to directly predict the text in a two-dimensional space instead of a one-dimensional sequence. Inspired by FCN~\cite{fcn}, a Character Attention Fully Convolutional Network (CA-FCN) is proposed to predict the characters at pixel level. Then the word, as well as the location of each character, can be obtained by a word formation module, as shown in Fig.~\ref{subfig-2:introduction}. In this way, the procedures of compressing and slicing the features, which are widely used in the sequence-based methods, are avoided. Profiting from the higher dimensional perspective, the proposed method is much more robust than the previous sequence-based methods in terms of text shapes, background noises, and imprecise localizations from the detection stage~\cite{LiaoSBWL17,LiaoSB18,liao2018rotation}. Character-level annotations are needed in our proposed method. However, the character annotations are free of labor because only public synthetic data is used in the training period, where the annotations are easy to obtain.

The contributions of this paper can be summarized as follows:
(1) A totally different perspective for recognizing scene text is proposed. Different from the recent works which treat the text recognition problem as a sequence recognition problem in one-dimensional space, we propose to solve the problem in two-dimensional space.
(2) We devise character attention FCN for scene text recognition. To the best of our knowledge, which can deal with images of arbitrary height and width, as well as naturally recognize text in various shapes, including but not limited to oriented and curve shapes.  
(3) The proposed method achieves state-of-the-art performance on regular datasets and outperforms the existing methods with a large margin on irregular datasets.
(4) We investigate the network's robustness to imprecise localization in the text detection phase for the first time. This problem is important in real-world applications but was previously ignored. Experiments show that the proposed method is more robust to imprecise localization (see 
. Ablation study).

\section{Related Work}
Traditionally, scene text recognition systems firstly detect each character, using binarization or sliding-window operation, then recognize these characters as a word. Binarization-based methods, such as Extremal Regions~\cite{DBLP:conf/eccv/NovikovaBKL12} and Niblack's adaptive binarization~\cite{DBLP:conf/iccv/BissaccoCNN13}, find character pixels after binarization. However, text in the natural scene image may have varying backgrounds, fonts, colors or uneven illumination and so on, which binarization based methods can hardly handle. Sliding window methods use multi-scale sliding window strategy to localize characters from the text image directly, such as Random Ferns~\cite{DBLP:conf/iccv/WangBB11}, Integer Programming~\cite{DBLP:conf/cvpr/SmithFL11} and Convolutional Neural Network (CNN)~\cite{DBLP:conf/eccv/JaderbergVZ14}. For the word recognition stage, common methods are integrating contextual information with character classification scores, such as Pictorial Structure models, Bayesian inference, and Conditional Random Field (CRF), which are employed in \cite{DBLP:conf/iccv/WangBB11,DBLP:journals/pami/WeinmanLH09,DBLP:conf/bmvc/MishraAJ12,DBLP:conf/cvpr/MishraAJ12,DBLP:conf/cvpr/ShiWXZGZ13}.

Inspired by speech recognition, recent works designed an encoder-decoder framework, where text in images are encoded into feature sequences and then decoded as characters. With the development of the deep neural network, convolutional features are extracted at encoder stage, and then RNN or CNN network is applied to decode these features, then CTC is used to form the final word. This framework was proposed by \cite{DBLP:journals/pami/ShiBY17}. Later they also developed an attention-based STN for rectifying text distortion, which is useful to recognize curved scene text~\cite{DBLP:conf/cvpr/ShiWLYB16}. Based on this framework, subsequent works~\cite{conf/aaai/HeH0LT16,DBLP:conf/icfhr/WuYLLW16,DBLP:conf/aaai/LiuCW18} also focus on irregular scene text.

The encoder-decoder framework has dominated current text recognition works. Many systems based on this framework have achieved state-of-the-art performance. However, text in scene images are distributed in a two-dimensional space, which is different from speech. The encoder-decoder framework just considers them as one-dimensional sequences, bringing some problems. For example, compressing a text image into a feature sequence may lose key information and add extra noise, especially when the text is curved or seriously distorted. 

There are some works that tried to improve some disadvantages of the encoder-decoder framework. \cite{bai2018edit} found that when considering the scene text recognition problem under the attention-based encoder-decoder framework, the misalignment between the ground truth strings and the attention's output sequences of the probability distribution, which is caused by missing or superfluous characters, will confuse and mislead the training process. To handle this problem, they propose a method called edit probability which considered losses including not only the probability distribution but also the possible occurrences of missing/superfluous characters. \cite{cheng2018aon} aimed to handle oriented text and realized that it is hard for the current encoder-decoder framework to capture the deep features of the oriented text. To solve this problem, they encode the input image to four feature sequences of four directions to extract scene text features in those directions. \cite{masktextspotter} proposed an instance segmentation model for word spotting, which uses an FCN-based method in its recognition part. However, it focused on the end-to-end word spotting task and no discussion is applied to verify the recognition part.

In this paper, we consider text recognition from the two-dimensional perspective and design a character attention FCN to deal with text recognition problem, which can naturally avoid those disadvantages of the encoder-decoder frameworks. 
For example, compressing a text image into a feature sequence may lose key information and add extra noise, especially when the text is curved or seriously distorted.
The proposed method obtains high accuracy on both regular and irregular text, Meanwhile, it is also robust to imprecise localization in the text detection phase.

\section{Methodology}
\subsection{Overview}
The whole architecture of our proposed method consists of two parts. The first part is a Character Attention FCN (CA-FCN) which predicts the characters at pixel level. Another part is a word formation module which groups and arranges the pixels to form the final word result.
\subsection{Character attention FCN}

The architecture of CA-FCN is basically a fully convolutional network, as shown in Fig.~\ref{fig:network}. We use VGG-16 as the backbone while dropping the fully connected layers and removing its pooling layers of stage-4 and stage-5. Besides, a pyramid-like structure~\cite{fpn} is adopted to handle varying scales of characters. The final output is of shape $\frac{H}{2} \times \frac{W}{2} \times C$, where $H$, $W$ are the height and width of the input image and $C$ is the number of classes including character categories and background. It can handle text of various shapes by predicting characters in a two-dimensional space.

\subsubsection{Character attention module}
Attention module plays an important role in our network. Natural scene text recognition suffers from complex backgrounds, shadow, irrelevant symbols and so on. Moreover, characters in natural images are usually crowded, which can hardly be separated. To deal with those problems, inspired by~\cite{res-attention}, we propose a character attention module to highlight the foreground characters and weaken the background, as well as separate adjacent characters, as illustrated in Fig.~\ref{fig:network}. Attention module is appended to each output layer of VGG16. The low-level attention models mainly focus on the appearance, such as edge, color, and texture. And the high-level modules can extract more semantic information. The character attention module can be expressed as follows:
\begin{equation}
F_o = F_i \otimes (1+A)
\end{equation}
where $F_i$ and $F_o$ are the input and output feature map respectively; $A$ indicates the attention map; $\otimes$ means element-wise multiplication. The attention map is generated by two convolutional layers and a two-class (characters and background) soft-max function where $0$ represents background and $1$ indicates characters. The attention map $A$ is broadcast to the same shape as $F_i$ to achieve element-wise multiplication. 
Compared with \cite{res-attention}, our character attention module uses a simpler network structure, profiting from the character supervision.
The effectiveness of the character attention module is discussed in Sec. Ablation study.

\begin{figure}[!htp]
\begin{center}
\captionsetup[subfigure]{justification=centering}
    \centering
\subfloat[\label{subfig-1:deform}]{%
       \includegraphics[width=0.2\textwidth]{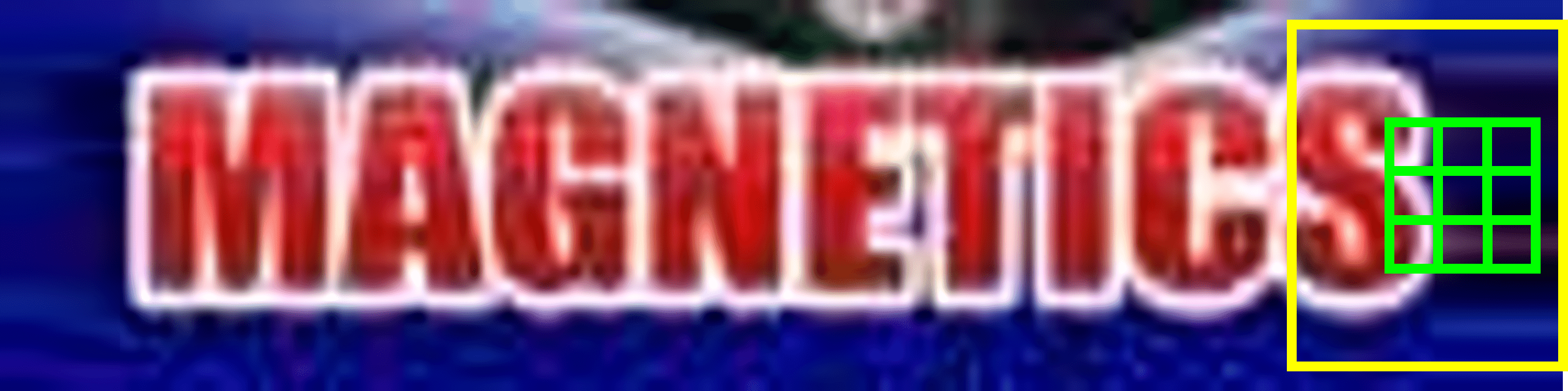}
     }
\subfloat[\label{subfig-2:deform}]{%
       \includegraphics[width=0.2\textwidth]{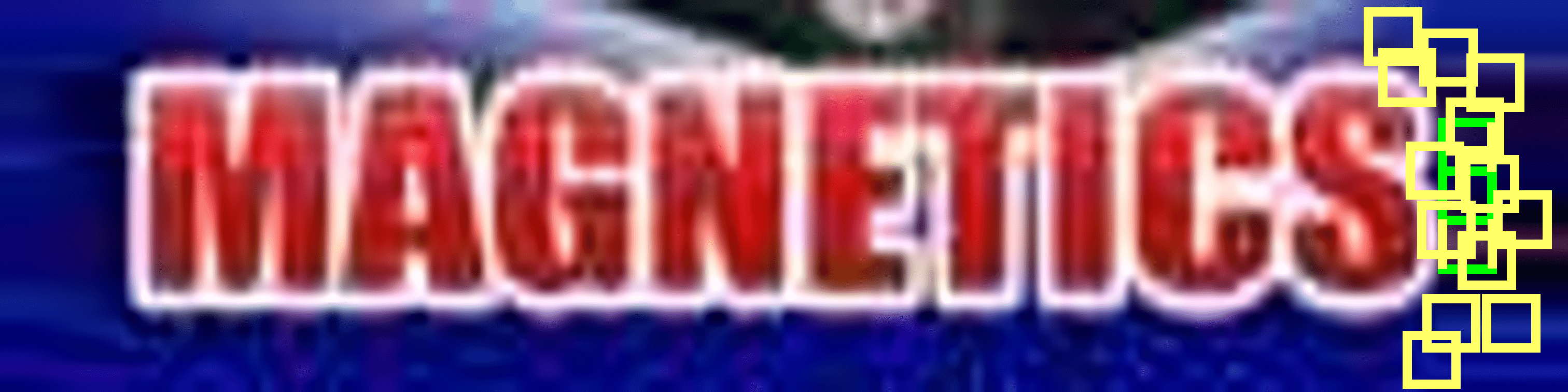}
     }
\end{center}
\caption{Illustration of our deformable convolution. (a) normal convolution; (b) deformable convolution with $3*1$ convolution. The green boxes indicate convolutional kernels. The yellow boxes mean the regions covered by receptive fields. The receptive fields out of the image are clipped.}
\label{fig:deform}
\end{figure}

\subsubsection{Deformable convolution}
As shown in Fig.~\ref{fig:network}, deformable convolution~\cite{deform} is applied in stage-4 and stage-5. The deformable convolution learns offsets of the convolution kernel, which provides more flexible receptive fields for the character prediction. The kernel size of deformable convolution is set to $3 \times 3$ as default. The kernel size of the convolution after the deformable convolution is set to $3 \times 1$. In Fig.~\ref{fig:deform}, there is a toy description of normal convolution, the deformable convolution with $3 \times 1$ convolutional kernel, as well as their receptive fields. The image in Fig.~\ref{fig:deform} is an expanded text image where more background is included in the image. Since most of the training images are cropped with tight bounding boxes, and the normal convolution contains a lot of character information due to the fixed receptive field, it tends to predict the extra background as a character. However, if deformable convolution and $3 \times 1$ convolution kernel are applied, with better and more flexible receptive field, the extra background can be predicted correctly. Note that the extra background is very common in real-world applications as the detection results may be inaccurate. Thus, the robustness on expanded text images is significant. The effectiveness of the deformable convolution is discussed in Sec. Ablation study by experiments.

\subsection{Training}
\paragraph{Label generation}
Let $b=(x_{min}, y_{min}, x_{max}, y_{max})$ be the original bounding boxes of characters, which can be expressed as the minimum axis-aligned rectangle boxes that covers the characters. The ground truth character regions $g=(x_{min}^g,y_{min}^g,x_{max}^g,y_{max}^g)$ can be calculated as follows:
\begin{equation} \label{label_gen}
\begin{split}
w & = x_{max} - x_{min} \\
h & = y_{max} - y_{min} \\
x_{min}^g & = (x_{min} + x_{max} - w \times r ) / 2 \\
y_{min}^g & = (y_{min} + y_{max} - h \times r ) / 2 \\
x_{max}^g & = (x_{min} + x_{max} + w \times r ) / 2 \\
y_{max}^g & = (y_{min} + y_{max} + h \times r ) / 2 
\end{split}
\end{equation}
where $r$ is the shrink ratio of the character regions. We shrink the character regions because the adjacent characters tend to be overlapped without shrinking. The shrink process can reduce the difficulty of the word formation. Specifically, we set $r$ to $0.5$ and $0.25$ for the attention supervision and the final output supervision respectively.

\begin{figure}[!hbp]
\begin{center}
\captionsetup[subfigure]{justification=centering}
    \centering
\subfloat[\label{subfig-1:labels}]{%
       \includegraphics[width=0.14\textwidth]{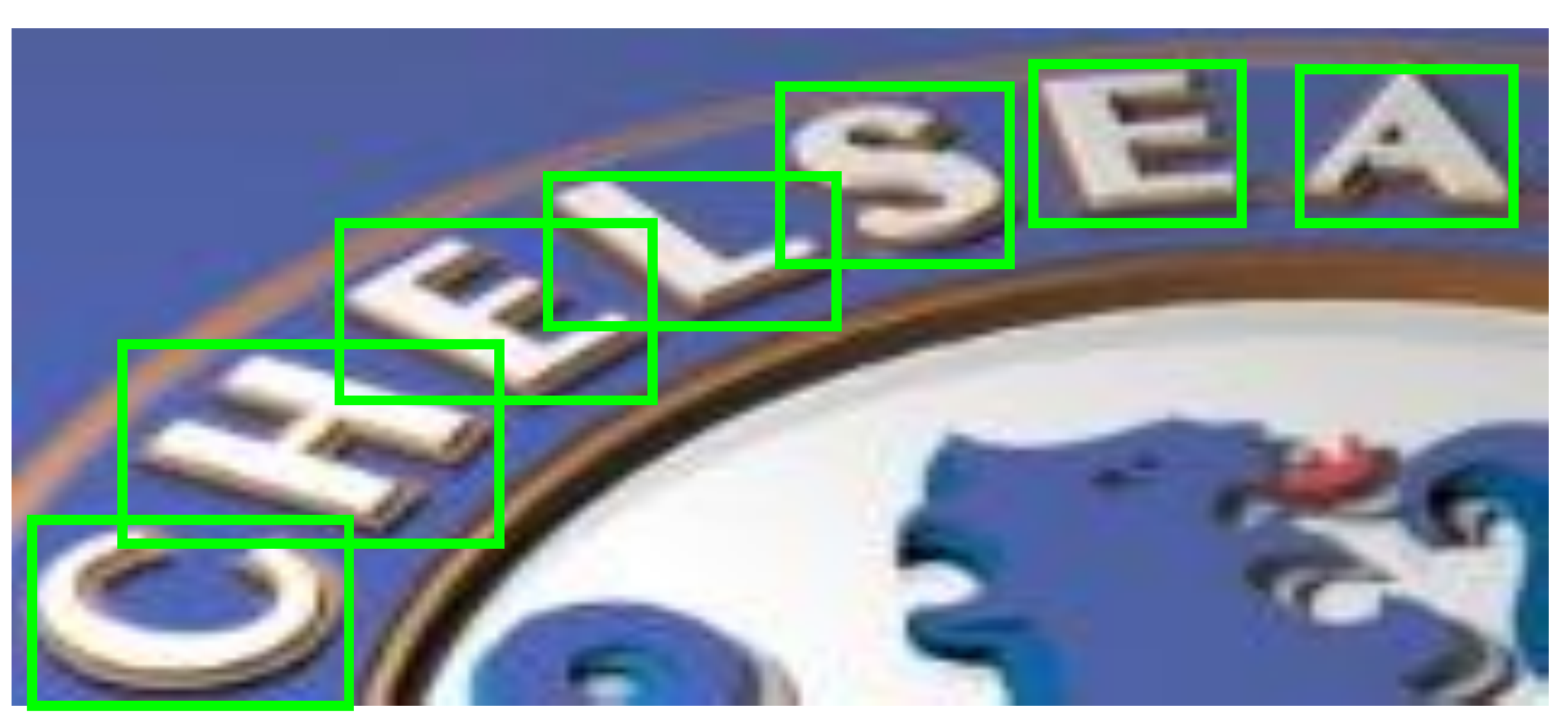}
     }
\subfloat[\label{subfig-2:labels}]{%
       \includegraphics[width=0.14\textwidth]{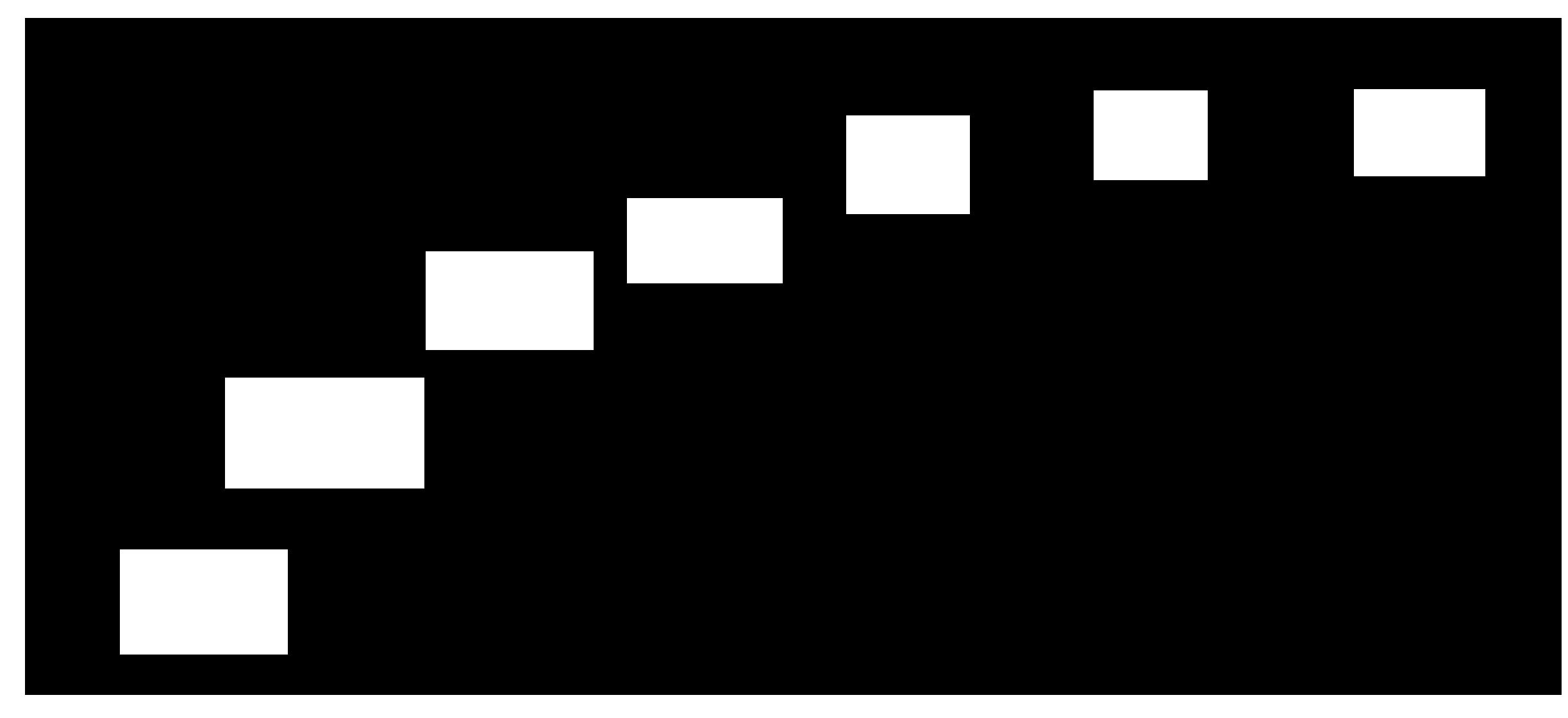}
     }
\subfloat[\label{subfig-3:labels}]{%
       \includegraphics[width=0.14\textwidth]{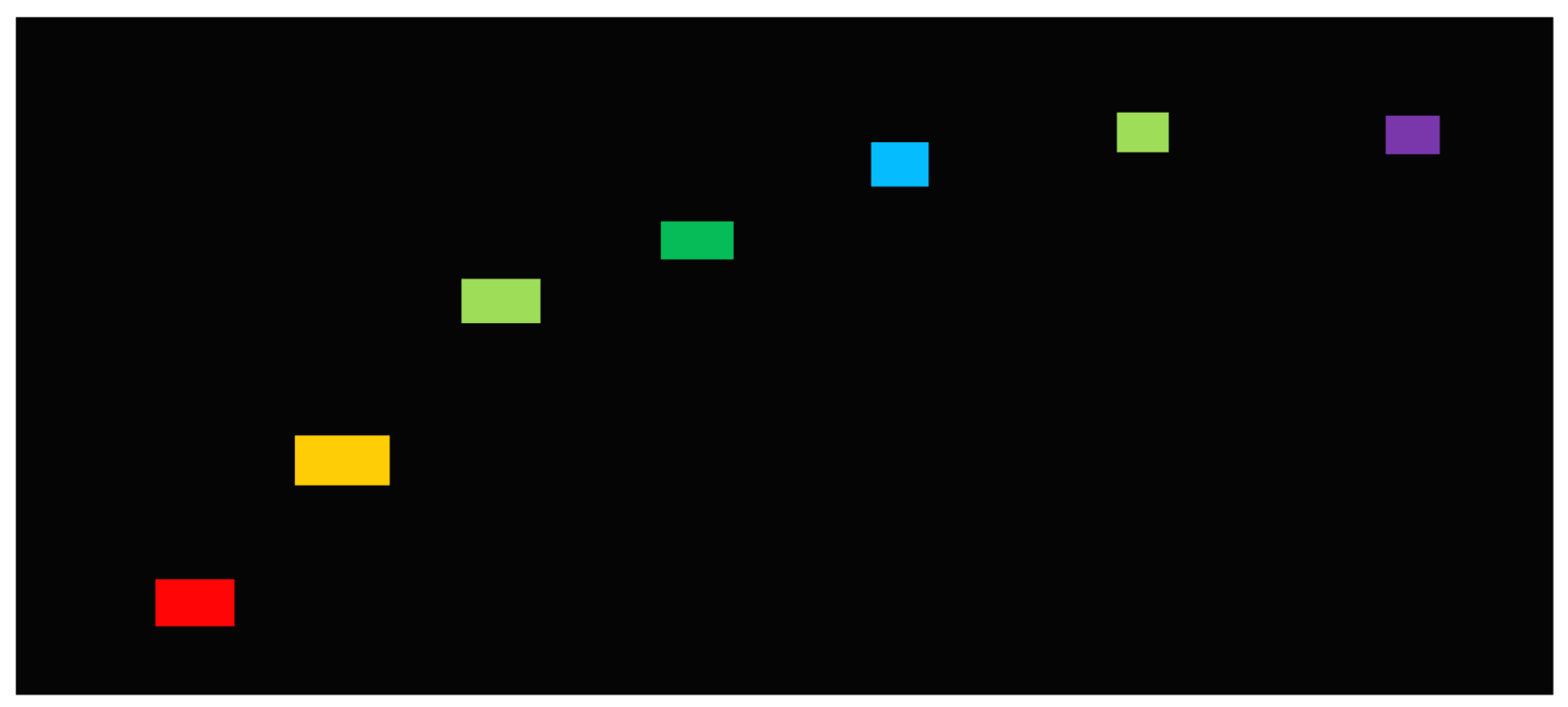}
     }
\end{center}
\caption{Illustration of ground truth generation. (a) Original bounding boxes; (b) Ground truth for character attention; (c) Ground truth for character prediction, where different colors represent different character classes.}
\label{fig:labels}
\end{figure}

\paragraph{Loss function}
The loss function is a weighted sum of the character prediction loss function $L_p$ and the character attention loss function $L_a$:
\begin{equation}
L = L_p + \alpha \sum_{s=2}^{5}L_a^s
\end{equation}
where $s$ indicates the index of the stages, as shown in Fig.~\ref{fig:network}; $\alpha$ is empirically set to $1.0$.

The final output of the CA-FCN is of shape $\frac{H}{2} \times \frac{W}{2} \times C$, where $H$, $W$ are the height and width of an input image respectively. $C$ is the number of classes including character classes and background. Assume that $X_{i,j,c}$ is one of the element of the output map, where $i \in \{1, ..., \frac{H}{2}\}$, $j \in \{1, ..., \frac{W}{2}\}$, and $c \in \{0,1, ..., C-1\}$; $Y_{i,j} \in \{0,1, ..., C-1\}$ indicates the corresponding class label. The prediction loss can be calculated as follows:
\begin{equation}
\begin{aligned}
L_p = & -\frac{4}{H \times W}\sum_{i=1}^{H/2}\sum_{j=1}^{W/2}W_{i,j}(\\
& \sum_{c=0}^{C-1} 
(Y_{i,j}==c) log(\frac{e^{X_{i,j,c}}}{\sum_{k=0}^{C-1} e^{X_{i,j,k}}})),
\end{aligned}
\end{equation}
where $W_{i,j}$ is the corresponding weight of each pixel. Assume that $N=\frac{H}{2} \times \frac{W}{2}$ and $N_{neg}$ is the number of background pixels. The weight can be calculated as follows:
\begin{equation}
  W_{i,j} = 
  \begin{cases}
    N_{neg} / (N - N_{neg})& \text{if } Y_{i,j}>0, \\
    1& \text{otherwise}
  \end{cases}
\end{equation}
The character attention loss function is a binary cross entropy loss function which take all characters labels as $1$, background label as $0$:
\begin{equation}
\begin{aligned}
L_a^s = & -\frac{4}{H_s \times W_s}\sum_{i=1}^{H_s/2}\sum_{j=1}^{W_s/2}(\\
& \sum_{c=0}^{1} (Y_{i,j}==c) log(\frac{e^{X_{i,j,c}}}{\sum_{k=0}^{1} e^{X_{i,j,k}}}),
\end{aligned}
\end{equation}
where $H_s$ and $W_s$ are the height and width of the feature map in the corresponding stage $s$ respectively.

\subsection{Word formation module}
\begin{figure*}[ht]
\centering
\includegraphics[width=0.85\linewidth]{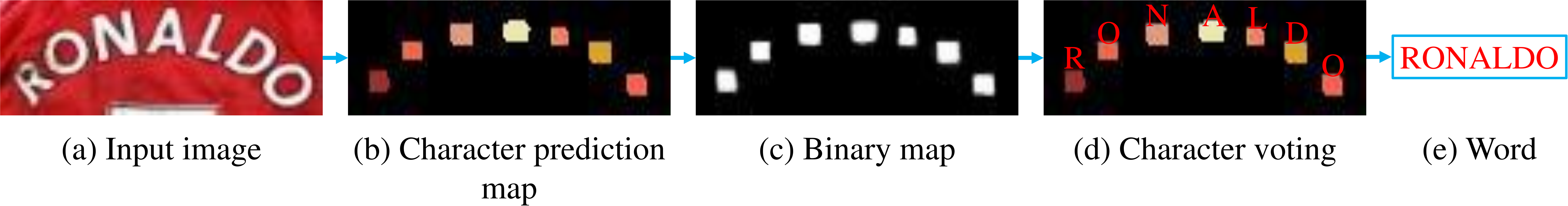}
\caption{Illustration of the word formation module.}
\label{fig:word_form}
\end{figure*}

The word formation module converts the accurate, two-dimensional character maps predicted by CA-FCN into character sequence. As shown in Fig.~\ref{fig:word_form}, we firstly transform the character prediction map into a binary map with a threshold to extract the corresponding character regions; then, we calculate the average values of each region for $C$ classes and assign the class with the largest average value to the corresponding region; finally, the word is formed by sorting the regions from left to right. In this way, both the word and location of each character are produced. The word formation module assumes that words are roughly sorted from left to right, which may not work in certain scenarios. However, if necessary, a learnable component can be plugged into CA-FCN. The word formation module is simple yet effective, with only one hyper-parameter (the threshold to form binary map), which is set to $240/255$ for all experiments.

\begin{figure*}[ht]
\centering
\includegraphics[width=0.85\linewidth]{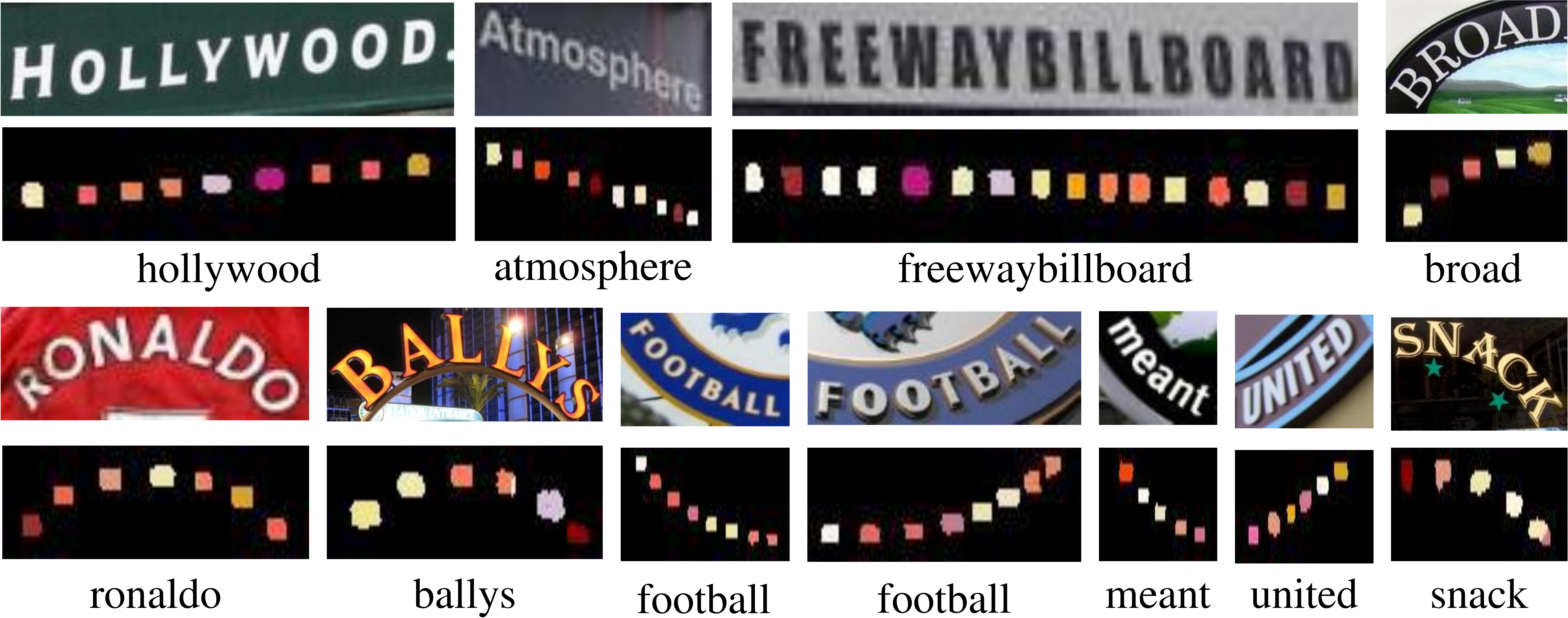}
\caption{Visualization of character prediction maps on IIIT and CUTE. The character prediction map generated by the CA-FCN is visualized with colors.}
\label{fig:visu}
\end{figure*}

\section{Experiments}
\subsection{Datasets}
Our proposed CA-FCN is purely trained on the synthetic datasets without real-world images. The trained model, without further fine-tuning,  was evaluated on 4 benchmarks including regular and irregular text datasets.

\textbf{SynthText} is a synthetic text dataset proposed in \cite{gupta2016synthetic}. It contains 800,000 training images which are aimed at text detection. We crop them based on their word bounding boxes. It generates about 7 million images for text recognition. These images are with character-level annotations.

\textbf{IIIT5k-Words} (IIIT)~\cite{DBLP:conf/cvpr/MishraAJ12} consists of 3000 test images collected from the web. It provides two lexicons for each image in the dataset, which contains 50 words and 1000 words respectively.

\textbf{Street View Text} (SVT)~\cite{DBLP:conf/iccv/WangBB11} comes from the Google Street View. The test set consists of 647 images. It is challenging due to its low resolution and noises. A 50-word lexicon is given for each image.

\textbf{ICDAR 2013} (IC13)~\cite{karatzas2013icdar} contains 1015 images and no lexicon is provided. We remove images that contain non-alphanumeric characters or have less than three characters, following previous works.

\textbf{CUTE}~\cite{cute} is a dataset consists of 288 images with a lot of curved text. It is challenging because the shapes vary hugely. No lexicon is provided.

\subsection{Implementation details}
\subsubsection{Training}
Since our network is fully convolutional, there is no restriction on the size of input images. We adopt multi-scale training to make our model more robust. The input images are randomly resized to $32 \times 128$, $48 \times 192$, and $64 \times 256$. Besides, data augmentation is also applied in the training period, including random rotation, hue, brightness, contrast, and blur. Specifically, we randomly rotate the image with an angle in the range of $[-15^\circ, 15^\circ]$.
We use Adam~\cite{adam} to optimize our training with the initial learning rate $10^{-4}$. The learning rate is decreased to $10^{-5}$ and $10^{-6}$ at epoch 3 and epoch 4. The model is totally trained for about 5 epochs. The number of character classes is set to $38$, including $26$ alphabet, $10$ digitals, $1$ special character which represents those characters out of alphabet and digitals, and $1$ background. 

\subsubsection{Testing}
At runtime, images are resized to $H_t \times W_t$, where $H_t$ is fixed to $64$ and $W_t$ is calculated as follows:
\begin{equation}
  W_t = 
  \begin{cases}
    W*H_t/H& \text{if } W/H>4, \\
    256& \text{otherwise}
  \end{cases}
\end{equation}
where $H$ and $W$ are the height and width of the origin images. 

The speed is about $45$ FPS on IC13 dataset with a batch size of 1, where the CA-FCN costs $0.018$ second per image and word formation module costs $0.004$ second per image on average. Higher speed can be achieved if the batch size increases. We test our method with a single Titan Xp GPU.

\begin{table*}[!ht]
\caption{Results across different methods and datasets. ``50'' and ``1k'' indicate the sizes of the lexicons. ``0'' means no lexicon. ``data'' indicates using extra synthetic data to fine-tune the model.}
\label{tab:performance}
\centering
\begin{tabular}{|l|ccc|cc|c|c|}
\hline 
\multirow{2}{*}{\textbf{Methods}}  & \multicolumn{3}{c|}{\textbf{IIIT}} & \multicolumn{2}{c|}{\textbf{SVT}}  & \textbf{IC13}  & \textbf{CUTE}\tabularnewline
\cline{2-8} 
& 50 & 1k & 0 & 50 & 0  & 0 & 0\tabularnewline
\hline 
\cite{DBLP:conf/iccv/WangBB11}  & - & - & - & 57.0 & -  & - & -\tabularnewline
\cite{DBLP:conf/bmvc/MishraAJ12}  & 64.1 & 57.5 & - & 73.2 & -  & - & -\tabularnewline
\cite{WangWCN12}  & - & - & - & 70.0 & -  & - & -\tabularnewline
\cite{AlmazanGFV14}  & 91.2 & 82.1 & - & 89.2 & - & - & -\tabularnewline
\cite{YaoBSL14}  & 80.2 & 69.3 & - & 75.9 & -  & - & -\tabularnewline
\cite{Rodriguez-Serrano15}  & 76.1 & 57.4 & - & 70.0 & - & - & -\tabularnewline
\cite{DBLP:conf/eccv/JaderbergVZ14}  & - & - & - & 86.1 & -  & - & -\tabularnewline
\cite{SuL14}  & - & - & - & 83.0 & - & - & -\tabularnewline
\cite{Gordo14}  & 93.3 & 86.6 & - & 91.8 & - & -  & -\tabularnewline
\cite{jaderberg2016reading}  & 97.1 & 92.7 & - & 95.4 & 80.7  & 90.8 & -\tabularnewline
\cite{JaderbergSVZ14b}  & 95.5 & 89.6 & - & 93.2 & 71.7  & 81.8 & -\tabularnewline
\cite{DBLP:journals/pami/ShiBY17} & 97.8 & 95.0 & 81.2 & 97.5 & 82.7  & 89.6 & -\tabularnewline
\cite{DBLP:conf/cvpr/ShiWLYB16}  & 96.2 & 93.8 & 81.9 & 95.5 & 81.9  & 88.6 & 59.2\tabularnewline
\cite{LeeO16}  & 96.8 & 94.4 & 78.4 & 96.3 & 80.7  & 90.0 & -\tabularnewline
\cite{DBLP:conf/nips/WangH17}  & 98.0 & 95.6 & 80.8 & 96.3 & 81.5  & - & -\tabularnewline
\cite{YangHZKG17} & 97.8 & 96.1 & - & 95.2 & - & - & 69.3\tabularnewline
\cite{ChengBXZPZ17}  & 99.3 & 97.5 & 87.4 & 97.1 & 85.9  & 93.3 & -\tabularnewline
\cite{cheng2018aon}  & 99.6 & 98.1 & 87.0 & 96.0 & 82.8  & - & 76.8\tabularnewline
\cite{bai2018edit}  & 99.5 & 97.9 & 88.3 & 96.6 & \textbf{87.5}  & \textbf{94.4} & -\tabularnewline
\hline 
Ours & \textbf{99.8} & \textbf{98.9} & \textbf{92.0} & 98.5 & 82.1  & 91.4 & 78.1 \tabularnewline
Ours+data  & \textbf{99.8}  & 98.8  & 91.9  & \textbf{98.8}  & 86.4  & 91.5 & \textbf{79.9} \tabularnewline
\hline 
\end{tabular}
\end{table*}

\subsection{Performances on benchmarks}
We evaluate our method on several benchmarks to indicate the superiority of the proposed method. Some results of IIIT and CUTE are visualized in Fig.~\ref{fig:visu}. As can be seen, our proposed method can handle various shapes of text. 

Quantitative results are listed in Tab.~\ref{tab:performance}. Compared to previous methods, our proposed method achieves state-of-the-art performance on most of those benchmarks. More specifically, ``Ours'' outperforms the previous state-of-the-art by $3.7$ percents on IIIT without lexicons. On irregular text dataset CUTE, $3.1$ percents improvement is achieved by ``Ours''. Note that no extra training data for curved text is included to achieve this performance. Comparable results are also performed on other datasets, including SVT, IC13. 

The training data of \cite{ChengBXZPZ17} consist of two synthetic datasets including Synth90k~\cite{synth90} and SynthText~\cite{gupta2016synthetic}. The former is generated according to a large lexicon which contains the lexicon of SVT and ICDAR, while the latter uses a normal corpus, where the distribution of words are not balanced. To fairly compared with \cite{ChengBXZPZ17}, we also generate extra 4 million synthetic images using the algorithm of SynthText with the lexicon used in Synth90k. As shown in Tab.~\ref{tab:performance}, after fine-tuning with the extra data, ``Ours+data'' also outperforms \cite{ChengBXZPZ17} on SVT.

\cite{bai2018edit} improves ~\cite{ChengBXZPZ17,DBLP:conf/cvpr/ShiWLYB16} by solving their misalignment problem and achieves excellent results in regular text recognition. However, it may fail in irregular text benchmarks such as CUTE due to its one-dimensional perspective. Moreover, we argue that our method can be further improved if the idea of \cite{bai2018edit} is well adapted to our word formulation module. Nevertheless, our method outperforms \cite{bai2018edit} on most of the benchmarks in Tab.~\ref{tab:performance}, especially on IIIT and CUTE. 

\cite{cheng2018aon} focuses on dealing with arbitrary-oriented text by introducing four one-dimensional feature sequences with different directions adaptively. Our method is more superior in recognizing the text of irregular shapes such as curve shape. As shown in Tab.~\ref{tab:performance}, our method outperforms \cite{cheng2018aon} on all benchmarks.

\begin{figure*}[!ht]
\centering
\includegraphics[width=0.9\linewidth]{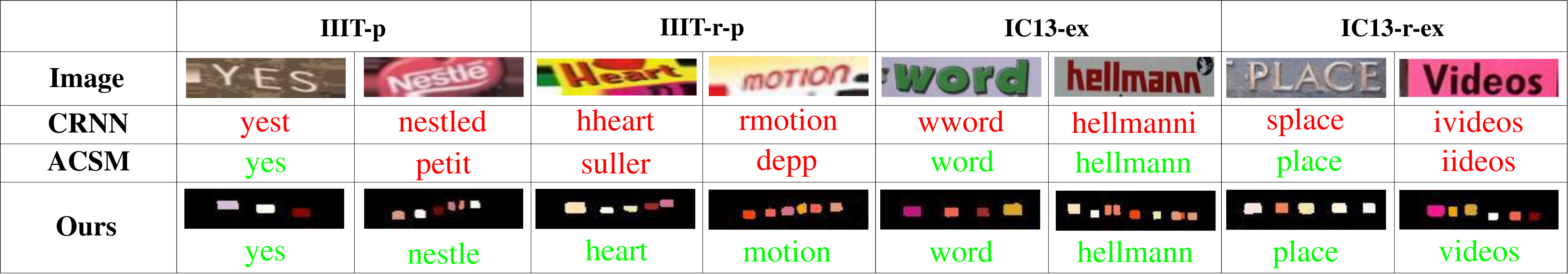}
\caption{Visualization of the character prediction maps on expanded datasets. Red: wrong results; Green: correct results.}
\label{fig:compare}
\end{figure*}

\begin{table*}[!ht]
\centering
\caption{Experimental results on expanded datasets. ``ac'': accuracy; ``gap'': the gap between the original dataset; ``ratio'' indicates the decreasing ratio compared to the accuracy on the original dataset.}
\label{tab:expand}
\scriptsize
\begin{tabular}{|c|c|c|c|c|c|c|c|c|c|c|c|c|c|c|}
\hline
\multirow{2}{*}{\textbf{Methods}} & \textbf{IIIT}          & \multicolumn{3}{c|}{\textbf{IIIT-p}}                    & \multicolumn{3}{c|}{\textbf{IIIT-r-p}}     & \textbf{IC13}          & \multicolumn{3}{c|}{\textbf{IC13-ex}}                   & \multicolumn{3}{c|}{\textbf{IC13-r-ex}}                 \\ \cline{2-15} 
                         & ac            & ac            & gap           & ratio          & ac        & gap       & ratio     & ac            & ac            & gap           & ratio          & ac            & gap           & ratio          \\ \hline
CRNN                    & 81.2          & 76.0              & -5.2              & 6.4\%               & 72.4          &  -8.8         & 10.8\%          & 89.6          & 81.9               & -7.7              & 8.6\%                &76.7               & -12.9              & 14.4\%               \\ \hline
ACSM                 & 85.4          & 79.1              & -6.3              & 7.4\%               & 74.9          &  -10.5         & 12.3\%          & 88.0         & 81.2              & -6.8             & 7.7\%                & 70.0              & -18.0              & 20.5\%               \\ \hline
baseline                  & 90.5 & 87.0 & -3.5 & 3.9\% & 85.7 & -4.8 & 5.3\% & 90.5 & 83.2 & -7.3 & 8.1\% & 82.3 & -8.2 & 9.1\% \\ \hline
baseline + attention                  & 91.0 & 86.7 & -4.3 & 4.7\% & 85.7 & -5.3 & 5.8\% & 90.1 & 85.6 & -4.5 & 5.0\% & 83.0 & -7.1 & 7.9\% \\ \hline
baseline + deform                  & 91.4 & 87.6 & -3.8 & 4.2\% & 86.7 & -4.7 & 5.1\% & 91.1 & \textbf{87.4} & \textbf{-3.7} & \textbf{4.1\%} & \textbf{84.2} & \textbf{-6.9} & \textbf{7.6\%} \\ \hline
baseline + attention + deform                  & \textbf{92.0} & \textbf{89.3} & \textbf{-2.7} & \textbf{2.9\%} & \textbf{87.6} & \textbf{-4.4} & \textbf{4.8\%} & \textbf{91.4} & 87.2 & -4.2 & 4.6\% & 83.8 & -7.6 & 8.3\% \\ \hline
\end{tabular}
\end{table*}

\subsection{Ablation study}\label{sec:expanding}
Scene text recognition is usually a following step of scene text detection, whose results may be not as accurate as expected. Thus, performances of text spotting systems in real-world applications are significantly affected by the robustness of text recognition algorithms on expanded images. We conduct experiments with expanded datasets to show the effect of text bounding box variance on recognition and prove the robustness of our method.

For the datasets which have the original background, such as IC13, we expand their bounding boxes and then crop them from the original images. If no extra background is provided like IIIT, padding by repeating the border pixels is applied to these images. The expanded datasets are described below:

\noindent\textbf{IIIT-p}
Padding the images in IIIT with extra $10\%$ height vertically and $10\%$ width horizontally by repeating the border pixels.

\noindent\textbf{IIIT-r-p}
Separately stretching the four vertexes of the images in IIIT with a random scale up to $20\%$ of height and width respectively; border pixels are repeated to fill the quadrilateral images; images are transformed back to axis-aligned rectangles.

\noindent\textbf{IC13-ex}
Expanding the bounding boxes of the images in IC13 to expanded rectangles with extra $10\%$ height and width before cropping.

\noindent\textbf{IC13-r-ex}
Expanding the bounding boxes of the images in IC13 randomly with a maximum $20\%$ of width and height to form expanded quadrilaterals; The pixels in axis-aligned circumscribed rectangles of those images are cropped.

We compare our method with two representative sequence-based models including CRNN~\cite{DBLP:journals/pami/ShiBY17} and Attention Convolutional Sequence Model (ACSM)~\cite{DBLP:journals/corr/gao}. 
The model of CRNN is provided by its authors and the model of \cite{DBLP:journals/corr/gao} is re-implemented by ourself with the same training data as ours.
Qualitative results of the three methods are visualized in Fig.~\ref{fig:compare}. As can be observed, the sequence-based models usually predict extra characters if the images are expanded while CA-FCN is stable and robust. 

The quantitative results are listed in Tab.~\ref{tab:expand}. Compared to the sequence-based models, our proposed method is more robust among these expanding datasets. For example, on IIIT-p dataset, the gap ratio of CRNN is $6.4\%$ while ours is only $2.6\%$. Note that even though our performances on the standard datasets are higher, the gaps of ours are still much smaller than CRNN. 
As shown in Tab.~\ref{tab:expand}, both the deformable module and the attention module can improve the performance and the former also contributes to the robustness of the model. It indicates the effectiveness of the deformable convolution and the character attention module.

The possible reasons that our method is more robust than previous sequence-based models on expanded images could be:
Sequence-based models are in one-dimensional perspective, which are hard to endure extra background because the background noises are easy to encode into the feature sequence. In contrast, our method predicts the characters in a two-dimensional space, where both the characters and the background are the targeted predicting objects. The extra background is less likely to mislead the prediction of the characters. 

\section{Conclusion}

In this paper, we have presented a method called Character Attention FCN (CA-FCN) for scene text recognition, which models the problem in a two-dimensional fashion. By performing character classification at each pixel location, the algorithm can effectively recognize irregular as well as regular text instances. Experiments show that the proposed model outperforms existing methods on datasets with regular and irregular text. We also analyzed the impact of imprecise text localization to the performances of text recognition algorithms and proved that our method is much more robust. For future research, we will make the word formation module learnable and build an end-to-end text spotting system.

\subsubsection*{Acknowledgments}
This work was supported by National Key R\&D Program of China No. 2018YFB1004600, NSFC 61733007, to Dr. Xiang Bai by the National Program for Support of Top-notch Young Professionals and the Program for HUST Academic Frontier Youth Team.

\bibliographystyle{aaai}
\bibliography{reference}
\end{document}